\documentclass[12pt]{article}

\usepackage{tocloft}
\usepackage{graphicx}
\usepackage{natbib}
\usepackage[protrusion=true,expansion=true]{microtype}
\usepackage{amsmath,amsthm,amsfonts,amssymb,mathrsfs,dsfont,mathtools}
\usepackage{booktabs}
\usepackage{url}
\usepackage{float}
\usepackage{color}
\usepackage{parskip}
\usepackage[explicit]{titlesec}
\usepackage{threeparttable}
\usepackage{setspace}
\usepackage[margin=0.85in]{geometry}
\usepackage[font = onehalfspacing]{caption}
\usepackage[usenames,dvipsnames]{xcolor}
\usepackage[hyperfootnotes=true]{hyperref}
\usepackage[labelfont=bf]{caption}
\usepackage{authblk}
\usepackage{caption}
\usepackage[titletoc,toc,title]{appendix}
\usepackage{enumerate}
\usepackage{soul}
\usepackage{subfig}
\usepackage{accents}
\usepackage{subfiles} 
\usepackage[capitalise, nameinlink]{cleveref}
\usepackage{dsfont}
\usepackage{enumitem}

\usepackage{changepage}
\usepackage[noend]{algpseudocode}
\usepackage{tocloft}
\usepackage[protrusion=true,expansion=true]{microtype}
\usepackage{amsmath,amsthm,amsfonts,amssymb,mathrsfs,dsfont,mathtools}
\usepackage{float}
\usepackage{color}
\usepackage{parskip}
\usepackage[explicit]{titlesec}
\usepackage{threeparttable}
\usepackage{setspace}
\usepackage[font = onehalfspacing]{caption}
\usepackage[usenames,dvipsnames]{xcolor}
\usepackage[hyperfootnotes=true]{hyperref}
\usepackage[labelfont=bf]{caption}
\usepackage{authblk}
\usepackage{enumerate}
\usepackage{soul}
\usepackage{subfig}
\usepackage{accents}
\usepackage[capitalise, nameinlink]{cleveref}
\usepackage{diagbox}
\usepackage{array,multirow}

\usepackage[utf8]{inputenc} 
\usepackage[T1]{fontenc}    
\usepackage{hyperref}       
\usepackage{url}            
\usepackage{booktabs}       
\usepackage{amsfonts}       
\usepackage{nicefrac}       
\usepackage{microtype}      
\usepackage{lipsum}		
\usepackage{graphicx}
\usepackage{natbib}
\usepackage{doi}
\usepackage{subfiles}

\usepackage{adjustbox}
\usepackage{multirow}
\usepackage{amsmath}
\usepackage[section]{placeins}
\usepackage{algorithm}
\usepackage{algpseudocode}
\usepackage{dsfont}
\usepackage{comment}

\DeclareUnicodeCharacter{2212}{-}

\newcommand{\rc}[1]{{\color{red}{#1}}}

\setstcolor{red} 

\numberwithin{equation}{section}

\hypersetup{colorlinks = true, citecolor = MidnightBlue, linkcolor = MidnightBlue, urlcolor = MidnightBlue}

\titleformat{\section}{\normalfont\large\bfseries}{\thesection}{1em}{#1}
\titleformat{\subsection}{\normalfont\normalsize\bfseries}{\thesubsection}{1em}{#1}
\titleformat{\subsubsection}{\normalfont\normalsize\itshape}{\thesubsubsection}{1em}{#1}
\titlespacing\section{0pt}{12pt plus 4pt minus 2pt}{6pt plus 2pt minus 2pt}
\titlespacing\subsection{0pt}{12pt plus 4pt minus 2pt}{3pt plus 2pt minus 3pt}
\titlespacing\subsubsection{0pt}{12pt plus 4pt minus 2pt}{0pt plus 2pt minus 3pt}

\def\boxit#1{\vbox{\hrule\hbox{\vrule\kern6pt
          \vbox{\kern6pt#1\kern6pt}\kern6pt\vrule}\hrule}}

\definecolor{orange}{rgb}{1,0.5,0}
\definecolor{MyDarkBlue}{rgb}{0,0.08,0.45}

\definecolor{MyDarkGreen}{RGB}{0,100,0}

\def\fredcomment#1{\vskip 2mm\boxit{\vskip 2mm{\color{MyDarkBlue}\bf#1}
{\color{MyDarkBlue}\bf -- Fred\vskip 2mm}}\vskip 2mm}

\newtheorem{corollary}{Corollary}[section]
\newtheorem{definition}{Definition}[section]

\newtheorem{remark}{Remark}[section]
\newtheorem{theorem}{Theorem}[section]

\def\boxit#1{\vbox{\hrule\hbox{\vrule\kern6pt
          \vbox{\kern6pt#1\kern6pt}\kern6pt\vrule}\hrule}}

\definecolor{orange}{rgb}{1,0.5,0}
\definecolor{MyDarkBlue}{rgb}{0,0.08,0.45}

\def\fredcomment#1{\vskip 2mm\boxit{\vskip 2mm{\color{MyDarkBlue}\bf#1}
{\color{MyDarkBlue}\bf -- Fred\vskip 2mm}}\vskip 2mm}

\begin{document}
\title{\Large \bfseries Catastrophic-risk-aware reinforcement learning with extreme-value-theory-based policy gradients\thanks{We thank the Natural Sciences and Engineering Research Council of Canada (Davar: FIN-ML NSERC CREATE program, Godin: RGPIN-2017-06837 and RGPIN-2024-04593, Garrido: RGPIN-2017-06643) for their financial support.}
 } 
\author[a]{Parisa Davar}
\author[a,b]{Fr\'ed\'eric Godin\thanks{Corresponding author. \vspace{0.2em} \newline
{\mbox{\hspace{0.47cm}} \it Email addresses:} 
\href{mailto:parisa.daavar@gmail.com}{parisa.daavar@gmail.com} (Parisa Davar),
\href{mailto:jose.garrido@concordia.ca}{jose.garrido@concordia.ca} (Jose Garrido),
\href{mailto:frederic.godin@concordia.ca}{frederic.godin@concordia.ca} (Fr\'ed\'eric Godin).  }}
\author[a,c]{Jose Garrido}
\affil[a]{{\small Concordia University, Department of Mathematics and Statistics, Montr\'eal, Canada}}
\affil[b]{{\small Quantact Laboratory, Centre de Recherches Math\'ematiques, Montr\'eal, Canada}}
\affil[c]{{\small Digital Insurance And Long term risk (DIALog) research chair, France}}

\vspace{-10pt}
\date{ 
June 28, 2024}


\maketitle 

%


\begin{abstract}
\vspace{-5pt}
	
This paper tackles the problem of mitigating catastrophic risk (which is risk with very low frequency but very high severity) in the context of a sequential decision making process. This problem is particularly challenging due to the scarcity of observations in the far tail of the distribution of cumulative costs (negative rewards). A policy gradient algorithm is developed, that we call POTPG. It is based on approximations of the tail risk derived from extreme value theory. Numerical experiments highlight the out-performance of our method over common benchmarks, relying on the empirical distribution. An application to financial risk management, more precisely to the dynamic hedging of a financial option, is presented.
	
\bigskip 



\noindent \textbf{Keywords:} Risk-aware reinforcement learning, catastrophic risk, extreme value theory, peaks-over-threshold (POT), hedging.
\end{abstract}

\medskip



\doublespacing

\setcounter{page}{1}
\pagenumbering{arabic}


\section{Introduction}\label{se:intro}

Reinforcement learning (RL) consists in a set of methods allowing to optimize sequential decision processes through interactions with an environment. In traditional RL \citep[see][]{sutton2018reinforcement}, the primary objective is to maximize expected rewards. However, a subset of RL techniques, referred to as risk-aware reinforcement learning, aim to take risk also into account (i.e.~departure from the expected case), see for example \cite{wu1999minimizing}, \cite{borkar2001sensitivity}, \cite{tamar2012policy}, \cite{la2013actor}, \cite{chow2018risk}, \cite{greenberg2022efficient} and \cite{vijayan2023policy}. Integrating risk mitigation within the RL framework is of paramount importance in several areas, as policies producing high expected rewards together with a high risk might be unacceptable in certain circumstances. Financial risk management is an example key area where risk-aware RL methods are developed, see for instance \cite{buehler2019deep}, \cite{carbonneau2021equal}, \cite{cao2023gamma} and \cite{wu2023robust} for a few examples.

The present work is concerned with problems involving the minimization of catastrophic risk in a sequential decision process, which represents outcomes that are very rare but of extreme magnitude. Since such extreme events can cause very undesirable outcomes depending on the area of application, such as financial ruin, health-impeding consequences or accidents, mitigating their impact is very important. Another example of application is the measurement of capital requirements in finance or insurance, which are based on the average of outcomes in the very worst-case scenarios; minimizing capital requirements for a financial institution is a key determinant of its probability, as capital is costly to hold. 

Here, extreme risk is quantified through risk measures, reflecting the far tail of the distribution of total costs incurred by the agent. In particular, we consider the special case of the conditional Value-at-Risk, CVaR$_\alpha$, which represents the average outcome among the worst possible set of scenarios with probability $1-\alpha$. The main motivation of the work is that CVaR with a very high confidence level $\alpha$ is very poorly approximated with the empirical distribution, due to the scarcity of observations in the far tail. Such paucity can be caused either by the lack of extreme observations, or the inability to generate a sufficient number of scenarios that include enough extreme data points in a reasonable time frame.\footnote{Importance sampling (IS) methods can sometimes help with this issue, if a scenario generator is used. However, suitably improving performance with IS requires knowing the direction in which to tilt risk driver (i.e.~states) distributions to produce outcomes. Such information is not necessarily known in the context of highly complex and non-linear dynamics (for instance the optimization of large financial portfolio with non-linear instruments such as exotic options) and methods alternative to IS would be required in such cases.}
In most acute cases, extreme outcomes might even be outside the data range, by not having materialized yet. The scarcity of tail observations can be exacerbated if the cost outcomes from the problem have fat-tailed distributions.

Our main contribution is to develop a policy gradient method for risk-aware RL problems that is tailor-made for cases where catastrophic-level risk must be minimized. We refer to our algorithm as POTPG, as it relies on the peaks-over-threshold (POT) approach of extreme value theory (EVT) that allows extrapolating the far tail behavior of a distribution through asymptotic approximations leveraging the distribution from large (but not extreme) outcomes. To the best of our knowledge, we are the first to incorporate EVT results within reinforcement learning algorithms to tackle general sequential decision problems; our work can be seen as an extension of \cite{troop2022best} that explores catastrophic risk minimization within the multi-armed bandits framework, but did not tackle the more general Markov decision problem setting.

The paper is divided as follows. Section \ref{se:problem} describes the risk-aware sequential decision making problem considered here, and provides a conventional policy gradient algorithm to tackle the problem. Section \ref{se:ourmethod} proposes our POT policy gradient (POTPG) approach, a modified policy gradient algorithm based on extreme value theory estimates of the tail risk of a distribution. This algorithm is tailor-made to tackle catastrophic-level risk minimization. Section \ref{se:simul} benchmarks the performance of POTPG against the conventional approach in a controlled environment, whereas Section \ref{se:hedging} assesses its performance in a financial risk management application, namely option hedging optimization. The paper concludes with some final remarks in Section \ref{se:conclusion}. The Python code to replicate the various numerical experiments of this paper is available at \href{https://github.com/parisadavar/EVT-policy-gradient-RL}{https://github.com/parisadavar/EVT-policy-gradient-RL}.


\section{A risk-aware reinforcement learning problem and policy gradients}\label{se:problem}

We herein consider the framework of Markov decision processes\footnote{This work generalizes to non-Markovian state transition dynamics in a straightforward way.} to represent sequential decision problems. Such problems are represented by a set of time steps $\mathcal{T}=\{0 ,\ldots, T\}$, a state space $\mathcal{S}$, an action space $\mathcal{A}$, a cost space $\mathcal{C}$ and a sequence of transition probabilities characterizing the joint distribution of the next-step reward and state, given the current state and action, namely $\mathbb{P}[S_{t+1} \leq s',C_{t+1} \leq c | A_t=a, S_t=s]$ for $s,s' \in \mathcal{S}$, $a \in \mathcal{A}$, $c \in \mathcal{C}$ and $t\in \mathcal{T} \backslash \{T\}$. 

Without loss of generality, deterministic policies $\pi : \mathcal{S} \rightarrow \mathcal{A}$ are considered in this work.
Such framework gives rises to random state-action-cost sequences of the form $S_0, A_0, C_1$, $S_1, A_1, C_2, \ldots$, $S_{T-1}, A_{T-1}, C_T$, where at any time point $t$ the agent takes action $A_t=\pi(S_t)$ when encountering state $S_t$, at a cost of $C_t$, then the next-stage state $S_{t+1}$ is drawn randomly from the probability measure $\mathbb{P}$.

\subsection{A risk-aware reinforcement learning problem}

The risk-aware reinforcement learning problem considered here\footnote{Other formulations of risk-aware RL problems exist, such as maximizing the expected rewards under some risk constraints \citep[see for instance][]{prashanth2022risk}, or using dynamic risk-measures leading to time-consistent dynamic programs \citep[see][]{marzban2021deep,coache2023conditionally}.} is to find the optimal policy minimizing the risk associated with the cumulative discounted cost: denoting costs as $C^{(\pi)}_t$ to highlight their dependence on policy $\pi$, the problem considered can be written as
\begin{equation}
\label{RLproblem}
    \underset{\pi}{\min} \, \rho \left( \sum^T_{t=1} \gamma^t C^{(\pi)}_t\right)
\end{equation}
for some discount factor $\gamma\in (0,1]$ and a risk measure $\rho$ mapping random variables into perceived risk. In the classic non-risk-aware case \citep[see for instance][]{sutton2018reinforcement}, the risk measure is the expectation operator, namely $\rho=\mathbb{E}$. However, more general risk measures can be used to depict preferences of risk-aware agents. Since this work is concerned with catastrophic risk mitigation, we consider the specific case of the CVaR risk measure \citep{rockafellar2002conditional} depicting tail risk and defined as
\begin{eqnarray*}\label{eq: cvar}
    \text{CVaR}_{\alpha}(X)  = \frac{1}{1-\alpha} \int_{\alpha}^{1} q_{\gamma}(X)d\gamma, \quad \text{with } \, q_{\alpha}(X) = \inf \{x\in\mathbb{R}| F(x) \geq \alpha\},
\end{eqnarray*}
where $F$ is the cumulative distribution function (CDF) of $X$ and $q_{\alpha}(X)$ is its quantile at level $\alpha$. In what follows we write $q_{\alpha} = q_{\alpha}(X)$. When $X$ is an absolutely continuous random variable, as in this work, then CVaR has the alternative representation $\text{CVaR}_{\alpha}(X) = \mathbb{E}[X | X \geq q_{\alpha}]$, which can be interpreted as the average outcome among the set of the $100(1-\alpha)\%$ worst-case scenarios. This work considers catastrophic risk minimization, and as such, we consider very high levels for $\alpha$, i.e.~$\alpha$ very close to one.

\subsection{A policy gradient solution approach}

A natural approach to solve \eqref{RLproblem} is policy gradient methods. Policies are first restricted to a set of parametric policies $\pi_\theta$ with parameter vector $\theta$. In that case, Problem \eqref{RLproblem} reduces to solving
\begin{equation}\label{policy_gradient}
    \theta^* = \underset{\theta}{\arg\max} \ J(\theta) \quad \text{with } \,\, J(\theta) = \text{CVaR}_{\alpha} \left( \sum^T_{t=1} \gamma^t C^{(\pi_\theta)}_t\right).
\end{equation}
A common solution approach to the above problem is to use batch stochastic gradient descent, which leads to a sequence of parameter vectors $\{ \theta^{(j)}\}_{j \geq 1}$ obtained through
\begin{equation*}
    \theta^{(j+1)} = \theta^{(j)} + \eta_j \widehat{\nabla J} (\theta^{(j)}),
\end{equation*}
with $\{ \eta_n\}$ representing the learning schedule and $\widehat{\nabla J} (\theta^{(j)})$ representing a suitable (stochastic) approximation of the gradient. Here we use the celebrated ADAM algorithm of \cite{kingma2014adam}, with a step size parameter of $0.01$ to determine learning rate sequences $\{ \eta^{(j)}\}_{j \geq 1}$.

To approximate the gradient, a (forward) finite difference approach is used here: for some small $\epsilon>0$,
\begin{eqnarray} \label{eq: finite_diff}
    \widehat{\frac{\partial J}{ \partial \theta_i} } (\theta^{(j)}) &\approx& \frac{ \widehat{J}(\theta^{(j)}+ \epsilon \mathds{1}_i)-\widehat{J}(\theta^{(j)})}{\epsilon},
    \\ \widehat{\nabla J} (\theta^{(j)}) &=& \left[\widehat{\frac{\partial J}{ \partial \theta_1} } (\theta^{(j)}) \ldots \widehat{\frac{\partial J}{ \partial \theta_p} } (\theta^{(j)})\right]^\top , \label{eq: finite_diff2}
\end{eqnarray}
with $p$ being the dimension of the parameter vector and $\mathds{1}_i$ being the dummy vector containing zeroes, except for its $i^{th}$ element that is equal to one. The objective function $\widehat{J}(\theta)$ is approximated by sampling $n$ independent copies $X_1,\ldots,X_n$ of the cumulative costs $X=\sum^T_{t=1} \gamma^t C^{(\pi_{\theta})}_t$ obtained from a Monte-Carlo simulation, if a simulator of the environment is available, or alternatively through the application of the policy $\pi_{\theta}$ to real data, either in an online or offline fashion.

The most natural approach to obtain the estimate of the objective function consists in assuming that the empirical distribution of cumulative costs obtained through the mini-batch is close to the true distribution. Such method is referred to as the \textit{sample averaging method} and relies on
\begin{equation}\label{eq: SA}
     \widehat{J} = \widehat {CVaR}_{\alpha}(X) = \frac{\sum_{i=1}^{n} X_{i}\mathds{1}_{\{ X_{i} \geq \hat{q}_{\alpha}\}}}{\sum_{j=1}^{n} \mathds{1}_{\{ X_{j} \geq \hat{q}_{\alpha}\}}},
\end{equation}
where, denoting by $X_{(1)},\ldots, X_{(n)}$ the order statistics of the sample (i.e.~the sample sorted in increasing order), $\hat {q}_{\alpha}$ is the empirical quantile given by:
\begin{equation} 
    \hat{q}_{\alpha} = \widehat{VaR}_{\alpha}(X) = \inf \{x\in\mathbb{R}| \hat F (x) \geq \alpha\} 
    = X_{ (\lceil {\alpha n} \rceil)},
\end{equation}
with $\widehat F$ being the empirical CDF of $X$ given by 
    $\widehat F (x) = \frac{1}{n} \sum_{s=1}^{n} \mathds{1}_{\{X_{s} \leq x\}}$.

Unfortunately, when $\alpha$ is high and very close to one, the scarcity of observations can make the sample averaging approach very unstable in estimating the objective function, a problem which is exacerbated if the distribution of $X$ is heavy-tailed. This justifies the development of the EVT-based estimator described in the next section.


\section{Integrating extreme value theory estimates into policy gradients}\label{se:ourmethod}

This section first discusses the construction of CVaR estimates based on the peaks--over--threshold (POT) approach rooted in extreme value theory (EVT).\footnote{Alternative methods also based on EVT such as that of \cite{bairakdar2024gtd} could also have been contemplated.} The POT approach is discussed more in-depth in \cite{coles2001introduction} or \cite{mcneil2015quantitative}. 
The procedure integrating such estimates into policy gradient approaches is subsequently detailed. 

\subsection{Estimation of CVaR with the peaks-over-threshold approach}\label{se:EVTCVaREstim}

A wide set of distributions satisfy the following condition.
\begin{definition}
    A CDF $F$ is said to be in the maximum domain of attraction of the generalized extreme value distribution (GEVD) $H_{\xi}$ with parameter $\xi$,\footnote{The CDF of the GEVD is given by $H_{\xi}(x)= \exp( -(1+\xi x)^{\frac{-1}{\xi}})$ if $\xi \neq 0$, or $H_{\xi}(x)=\exp(-e^{-x}),$ if $\xi = 0$, with support $\{x : 1+\xi x >0\}$. } denoted $F \in MDA(H_{\xi})$, if there exist a sequence a positive numbers $\{a_n\}_{n \in \mathbb{N}}$ and a sequence of real numbers $\{b_n\}_{n \in \mathbb{N}}$, such that \begin{equation}\label{eq: fisher}
    \lim_{n \to \infty} F^{n}(a_{n}x + b_n) = H_{\xi}(x).
    \end{equation}
\end{definition}

Note that $F^n$ is the CDF of the maximum of $n$ i.i.d.~copies of a random variable with CDF $F$.

The $F \in MDA(H_{\xi})$ property characterizes the asymptotic behavior of distribution $F$. Indeed, define $F_u$, the distribution of excesses above threshold $u$, as
\begin{equation} \label{eq: excess equation}
    F_u(y) = P(X - u \leq y | X > u) = P(X \leq y + u | X > u) =\frac{F(y + u) - F(u)}{1 - F(u)}.
\end{equation}
Define also the generalized Pareto distribution (GPD) as follows.
\begin{definition}\label{def: GPD} The GPD with scale parameter $\sigma$ and shape parameter $\xi$ has a CDF
\begin{equation} \label{eq: GPD_cdf}
    G_{\xi,\sigma}(x) = \begin{cases} 1-(1+\frac{\xi x}{\sigma})^{\frac{-1}{\xi}}, & \mbox{if } \xi \neq 0, \\ 1 - e^{\frac{-x}{\sigma}}, & \mbox{if } \xi = 0, \end{cases}
\end{equation}
where the support is $x\geq 0$, for $\xi \geq 0$, and $0 \leq x \leq -\frac{\sigma}{\xi}$, for $\xi \leq 0$, and a probability density function (PDF)
\begin{equation} \label{eq: GPD_pdf}
    g_{\xi,\sigma}(x) = \begin{cases} \frac{1}{\sigma}(1+\frac{\xi x}{\sigma})^{\frac{-1}{\xi}-1}, & \mbox{if } \xi \neq 0, \\ \frac{1}{\sigma}e^{\frac{-x}{\sigma}}, & \mbox{if } \xi = 0. \end{cases}
\end{equation}
\end{definition}

Then the following result from \cite{balkema1974residual} or \cite{pickands1975statistical} states that when $F \in MDA(H_{\xi})$, the excess distribution $F_u$ is well-approximated asymptotically by a GPD distribution when $u$ is near to the essential supremum of distribution $F$. 

\begin{theorem}[Pickands–Balkema–de Haan]\label{theorem: Pickands}
If $F \in MDA(H_{\xi})$, there exists a positive measurable function $\sigma (u)$ such that 
\begin{equation}\label{eq: Pickands}
    \lim_{u\to y_0}\sup_{y_0\in [0, y_0 - u]} | F_u(y) - G_{\xi,\sigma (u)}(y)|,
\end{equation}
where $y_0 = sup\{ y \in \mathbb R; F(y)<1\} \leq \infty$ and $G_{\xi,\sigma (u)}(y)$. 
\end{theorem}

As described in Section 7.2 of \cite{mcneil2015quantitative}, such a result allows defining the following approximation for the CVaR of the variable $X$ with CDF $F$, which is based on the assumption that $F_u(y) \approx G_{\xi,\sigma (u)}(y)$ for $y>u$, i.e.~if $u$ is large enough.\footnote{Note that the condition $\xi \leq 1$ is required for the CVaR to exist, 
otherwise the GPD distribution has an infinite expectation.}

\begin{corollary}\label{col: EVT_approximation}
    Assume that $F \in MDA(H_{\xi})$ for some $\xi \in [0,1)$, and that $q_\alpha>u$. Let $\sigma = \sigma(u)$ satisfy conditions of Theorem \ref{theorem: Pickands}. Then for $s_{u,\alpha} = \frac{1-F(u)}{1-\alpha}$,
\begin{equation} \label{eq: cvar EVT approximation}
    {CVaR}_{\alpha}(X) \approx c_{u,\alpha} = \begin{cases} u + \frac{\sigma}{1-\xi}(1 + \frac{s^{\xi}_{u,\alpha}-1}{\xi}), & \mbox{if } \xi \neq 0, \\ u + \sigma (\log s_{u,\alpha} +1 ), & \mbox{if } \xi = 0. \end{cases}
\end{equation}
\end{corollary}

This points toward the following procedure, called the \textit{peaks-over-threshold} approach to estimate ${CVaR}_{\alpha}(X)$ based on a sample of i.i.d.~copies $X_1, \ldots , X_n$ of $X$:
\begin{enumerate}[noitemsep,nolistsep]
    \item Select a proper threshold $u$.
    \item Calculate sample values of excesses over threshold $u$, denoted $Y_1, \ldots, Y_k$ and defined as \mbox{$Y_i = X_{(n + 1-i)} - u$}, where $k$ is the number of sample observations $X_i$ above $u$.
    \item Fit a GPD distribution to the sample $Y_1, \ldots, Y_k$ to get estimates $(\hat\xi,\hat\sigma)$.
    \item Replace $(\xi,\sigma)$ and $F(u)$ with respective estimates $(\hat\xi,\hat\sigma)$ and $\hat{F}(u)$ into \eqref{eq: cvar EVT approximation} to get an approximation for ${CVaR}_{\alpha}(X)$.
\end{enumerate}
Since for any fixed $u$, excesses $Y_1, \ldots, Y_k$ are independent, Step 3 can be performed through maximum likelihood\footnote{De-biasing procedures could additionally be applied to adjust maximum likelihood estimates, such as in \cite{troop2021bias}.} by solving numerically
\begin{equation} \label{eq: MLEE}
    (\hat{\xi}, \hat{ \sigma}) = \underset{\xi, \sigma}{\arg\max}\sum_{i=1}^{k}  \ln g_{\xi, \sigma}(Y_i).
\end{equation}

Alternatively, a method-of-moments (MOM) estimator matching the first two moments\footnote{Here the MOM estimator requires that $\xi<1/2$ to ensure that the variance of the GPD be finite.} of the GPD distribution with those of the empirical distribution of excesses would lead to\footnote{This is because if $Y \sim \text{GPD}(\xi,\sigma)$, then $\mathbb{E}(Y) = \frac{\sigma}{1-\xi}$ if $\xi<1$ and $\text{Var}(Y) = \frac{\sigma^2}{(1-\xi)^2 (1-2\xi)}$ if $\xi<1/2$. Estimators in \eqref{MM} are obtained by equating $\mathbb{E}(Y)$ and $\text{Var}(Y)$ with $\bar{Y}$ and $S^2$, respectively.}
\begin{equation} \label{MM}
    \hat\xi = \frac{S^2 - \bar{Y}^2}{2S^2}, \quad \hat\sigma = \bar{Y} \left(\frac{S^2 + \bar{Y}^2}{2S^2}\right),
\end{equation} 
with $\bar{Y} = \frac{1}{k} \sum^k_{j=1} Y_j$ and $S^2 = \frac{1}{k} \sum^k_{j=1} (Y_j - \bar{Y})^2$.

The task in Step $1$, namely the selection of a suitable choice of threshold $u$ is challenging, as it entails seeking a proper bias-variance trade-off. Indeed, if $u$ is too low, the distribution tail behavior might not be well-approximated by its asymptotic GPD distribution, leading to high bias. Conversely, choosing a $u$ that is too large will imply a low number of excesses, which will lead to high variance for the GDP parameter estimators. A common approach  in the literature  is to manually select $u$ through visual inspection of the so-called Hill plot \citep[see][]{mcneil2015quantitative}. However, such a method is not appropriate in our setup since the choice of threshold $u$ needs to be repeated a very large number of times through the learning phase. As such, we rely on the \cite{bader2018automated} algorithm that performs automated selection of the threshold based on a sequence of Anderson-Darling goodness-of-fit tests. Such a procedure tests for a set of candidate values $u^{(1)} <\ldots<u^{(\ell)}$, and the smallest among these is selected as the threshold, which leads to a proper fit of the GPD to excesses over $u$. The implementation from \cite{troop2021bias} of the procedure is considered here and is detailed in Appendix \ref{App:Bader}. This modifications allows stabilizing estimates, for instance by not allowing estimated values of $\xi$ too close to one (to avoid the CVaR estimate exploding) and by using the sample averaging estimator as fallback, when none of the thresholds lead to a satisfactory fit of the GPD.


\subsection{Our proposed EVT policy gradient algorithm}\label{se:POTPG}

The POT-based CVaR estimation method from Section \ref{se:EVTCVaREstim}
is now integrated into the policy gradient estimation formula in \eqref{eq: finite_diff} to obtain a complete policy gradient learning procedure for the policy parameters $\theta$. This procedure, which we call the \textit{POTPG} algorithm (standing for for peaks-over-threshold policy gradient), is summarized in the \mbox{Algorithm \ref{alg:proposed_model_simulation}} box below.

\begin{algorithm}[H]
    \caption{POTPG}
    \label{alg:proposed_model_simulation}
    \begin{algorithmic} 
        \Require $\epsilon$ \text{(finite difference step)}, n \text{(number of episodes)}, M \text{(number of iterations)}

        \State Initialize $\theta_0$ through random sampling 
        \For{$j=0,\ldots,M-1$ } 
            \State Sample $n$ episodes of the MDP with policy $\pi_{\theta_j}$, and denote by $X_i = \sum^T_{t=1} \gamma^t C^{(\pi_{\theta^{(j)}})}_t$ the total discounted costs for the $i^{th}$ episode,
            
            \State Based on sample $X_1,\ldots, X_n$, obtain the estimates $\hat\xi$, $\hat\sigma$ and $\hat{F}(u)$ where the automated threshold selection method of Appendix \ref{App:Bader} is applied to determine $u$,
            
            \State Obtain the EVT-based estimate of $\widehat{J}(\theta^{(j)})$ through  \eqref{eq: cvar EVT approximation},
            
            \For{$i=1,\ldots,p$ }
                \State Sample $n$ episodes of the MDP with policy $\pi_{\theta^{(j)} + \epsilon \mathds{1}_{i}}$, and denote by $X_i = \sum^T_{t=1} \gamma^t C^{(\pi_{\theta^{(j)} + \epsilon \mathds{1}_{i}})}_t$ the total discounted costs for the $i^{th}$ episode,
                
                \State Based on sample $X_1,\ldots, X_n$ and threshold $u$, obtain the estimates $\hat\xi$, $\hat\sigma$ and $\hat{F}(u)$.
                
                \State Obtain the EVT-based estimate of $\widehat{J}(\theta^{(j)} + \epsilon \mathds{1}_{i})$ through  \eqref{eq: cvar EVT approximation},
            \EndFor
            \State Estimate the gradient $\widehat{\nabla J} (\theta^{(j)})$ through the finite difference scheme \eqref{eq: finite_diff}-\eqref{eq: finite_diff2},
            \State $\theta^{(j+1)} \leftarrow \theta^{(j)} + \eta_j \widehat{\nabla J} (\theta^{(j)})$, with $\eta_j$ as determined by the ADAM algorithm.
        \EndFor
        Return $\theta^{(M)}$
    \end{algorithmic}
\end{algorithm}

If a simulator of the environment is available, it can be desirable, within a given iteration $j$, to use the same random seed to perform the simulation of episodes under policy $\pi_{\theta^{(j)}}$ and these under policies $\pi_{\theta^{(j)} + \epsilon \mathds{1}_{i}}$, $i=1,\ldots,p$. This approach offers the advantage of isolating the impact of the policy alteration (from $\pi_{\theta^{(j)}}$ to $\pi_{\theta^{(j)} + \epsilon \mathds{1}_{i}}$) from the randomness associated with the generation of episodes; the latter can add noise to the gradient estimate. The same seed is used throughout all the experiments presented here to simulate episodes under the original and shocked policies.

Note also that we propose to use the same threshold $u$ to estimate $\widehat{J}(\theta^{(j)})$ and all $\widehat{J}(\theta^{(j)} + \epsilon \mathds{1}_{i})$ in the POTPG algorithm to enhance the stability in the gradient estimation.


\section{Simulation experiments in a controlled environment}\label{se:simul}

Several simulation experiments in a controlled environment are first conducted to assess the performance of the POTPG algorithm from Section \ref{se:POTPG} and compare it to the conventional sample averaging (SA) benchmark based on \eqref{eq: SA}. A simple simulation setting is considered to establish a proof of concept and highlight the potential usefulness of the POTPG algorithm. 
In such setting, we consider a single-dimension policy vector $\theta$ (i.e. $p=1$), and we assume the cumulative discounted cost is distributed according to a given family of distributions whose parameters depend on $\theta$.

More precisely,  assume that under policy $\pi_\theta$, $X\sim$ GPD$(\xi, \varsigma = (\theta-\vartheta)^2+b )$ for some $\xi \in (0,1)$, $b>0$ and $\vartheta \in \mathbb{R}$.\footnote{Here, to avoid confusion, we use $\varsigma$ instead of $\sigma$ to represent the scale parameter of the whole distribution instead of that of the tail.} Fix $\vartheta = 0.4$ and $b = 2$ and consider values $\xi=0.4$, $0.6$ or $0.8$ in subsequent experiments. Note that if $X \sim GPD(\xi, \varsigma)$, then the conditional exceedance $X-u | X>u \sim GPD(\xi, \varsigma + \xi u )$, meaning that the excess distribution of a GPD random variable is a GPD with the same shape parameter, and a scaling parameter that grows linearly with the threshold $u$. In that case, representing the tail distribution with a GPD is exact and not merely an asymptotic approximation. Such setting is used to test the POTPG algorithm in an ideal case with no misspecification of the tail distribution. The CVaR of the distribution with policy $\pi_\theta$ is then
\begin{equation*}
    \text{CVaR}_\alpha(X) = q_\alpha + \frac{ \varsigma + \xi q_\alpha}{1-\xi}
    = \frac{\varsigma}{1-\xi} \left(1 + \frac{\left(1-\alpha\right)^{-\xi}-1}{\xi} \right),
\end{equation*}
since $q_\alpha =\frac{\varsigma}{\xi}\left((1-\alpha)^{-\xi}-1\right)$. Therefore,
\begin{equation*}
    \theta^* = \underset{\theta}{\arg\min} \,\text{CVaR}_\alpha(X) = \vartheta, \quad \underset{\theta}{\min} \, \text{CVaR}_\alpha(X) = \frac{b}{1-\xi} \left(1 + \frac{\left(1-\alpha\right)^{-\xi}-1}{\xi} \right),
\end{equation*}
i.e.~the optimal policy is to set $\theta=\vartheta$ to minimize the scale parameter of the cumulative discounted costs.

In each simulation run, we consider $M=500$ iterations, in each of which $n=2,\!000$ cumulative discounted cost realizations are generated. A total of $R=50$ runs are performed. The finite difference step size for the gradient computation is $\epsilon = 0.01$. We set the initial policy parameter to $\theta^{(0)}=1$.
Define $\theta^{(j,r)}$ and $\widehat{J}^{(j,r)}$  respectively as estimates of the policy parameter and the objective function (CVaR) estimate  on the $j^{th}$ iteration of run $r$. 
We report the root-mean-square-error (RMSE) across the various runs for each iteration of policy parameters and the objective function associated as:
\begin{equation*}
    \text{RMSE}_\theta = \sqrt{\frac{1}{R} \sum^R_{r=1} \left( \theta^{(j,r)} - \theta^*\right)^2}, \quad
    \text{RMSE}_{\widehat{J}} = \sqrt{\frac{1}{R} \sum^R_{r=1} \left( \widehat{J}^{(j,r)} - J(\theta^*)\right)^2}.
\end{equation*}
The CVaR level $\alpha=0.998$ is chosen to depict catastrophic risk levels.

\begin{figure}%
\centering
\subfloat[$\text{RMSE}_\theta$, $\xi=0.4$]{\includegraphics[width=0.5\textwidth]{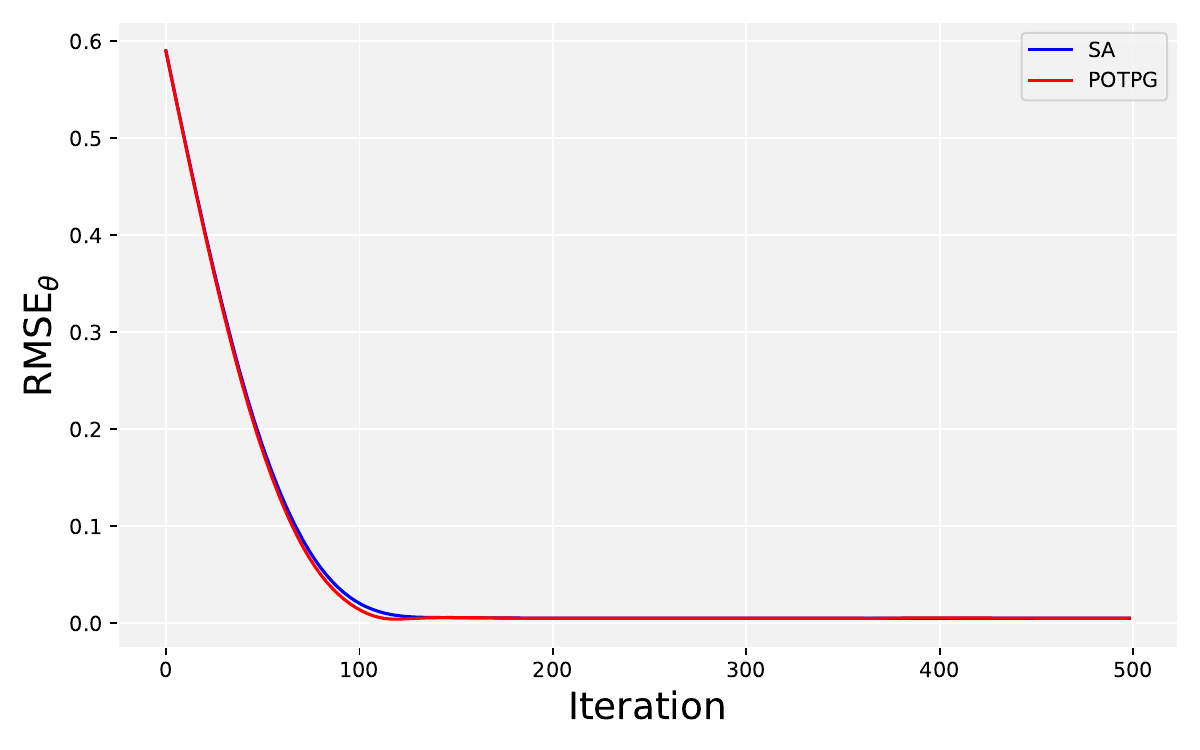}} 
\subfloat[$\text{RMSE}_{\widehat{J}}$, $\xi=0.4$]{\includegraphics[width=0.5\textwidth]{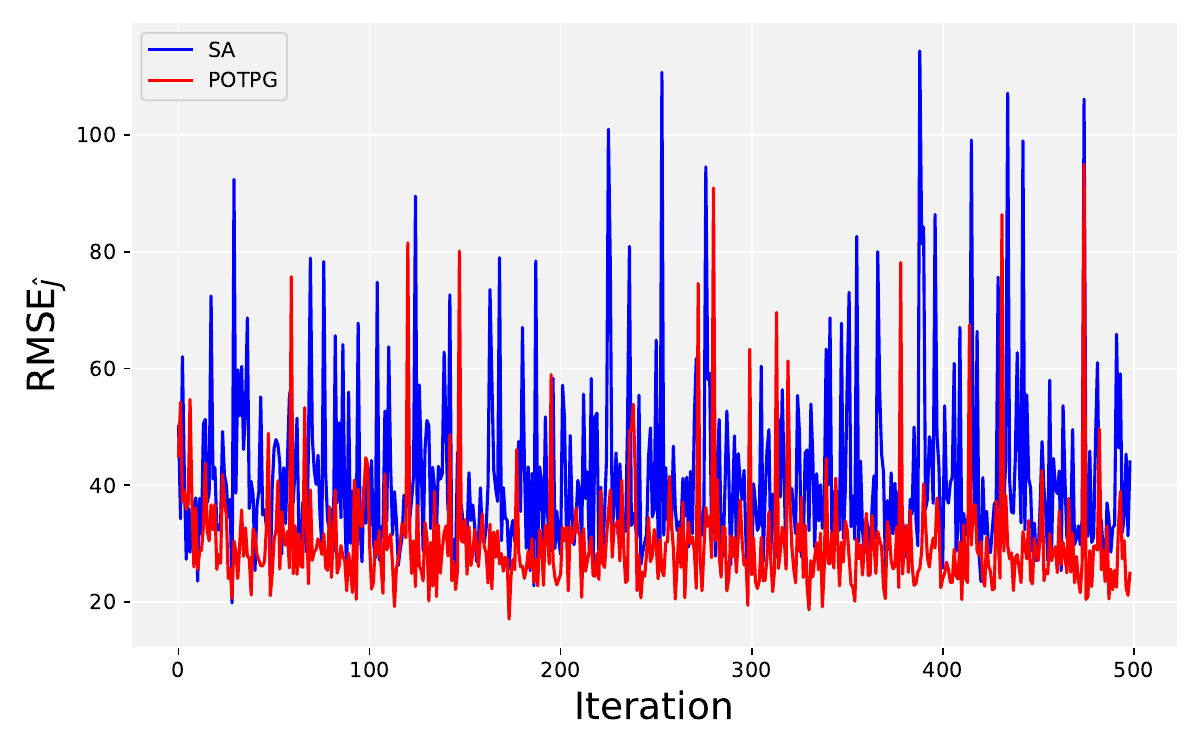}} \\ 
\subfloat[$\text{RMSE}_\theta$, $\xi=0.6$]{\includegraphics[width=0.5\textwidth]{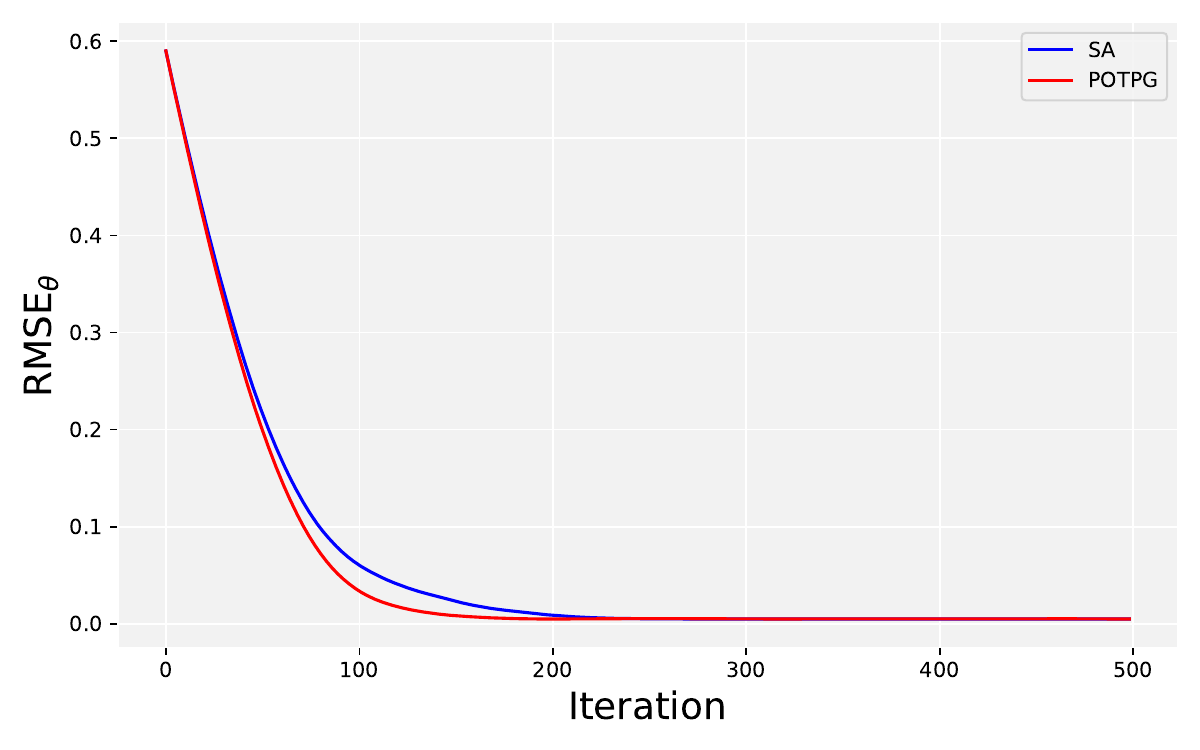}} 
\subfloat[$\text{RMSE}_{\widehat{J}}$, $\xi=0.6$]{\includegraphics[width=0.5\textwidth]{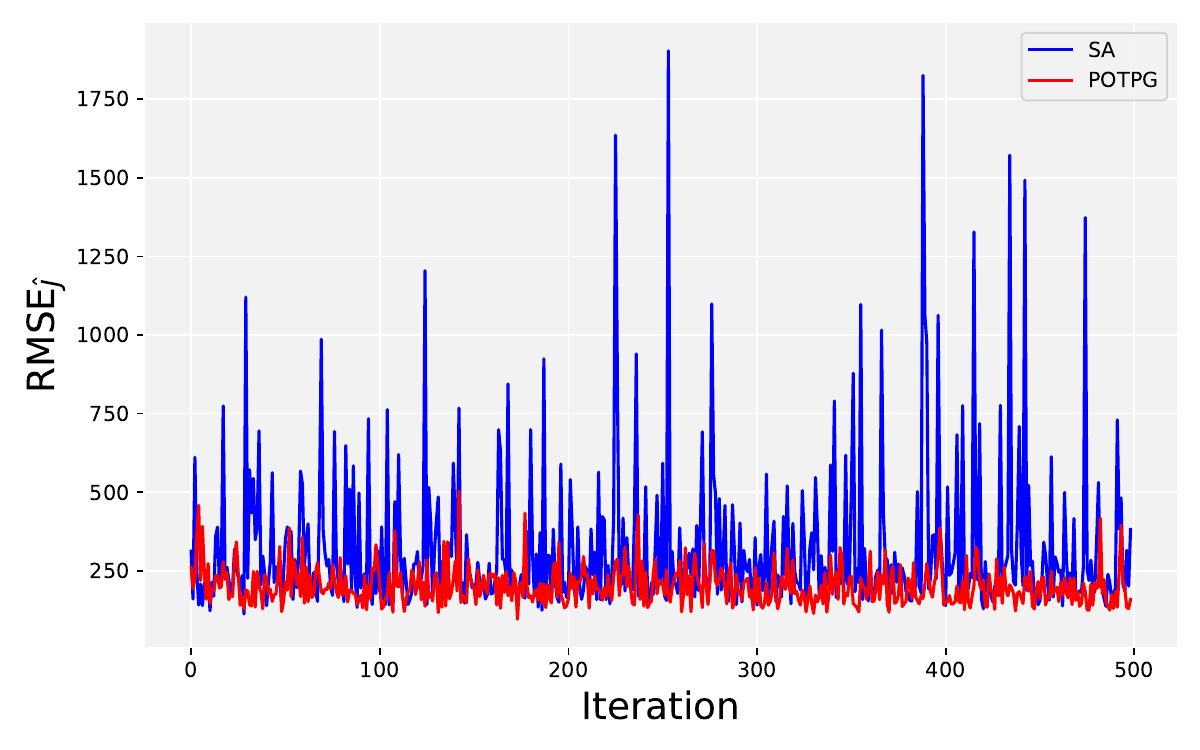}} \\ 
\subfloat[$\text{RMSE}_\theta$, $\xi=0.8$]{\includegraphics[width=0.5\textwidth]{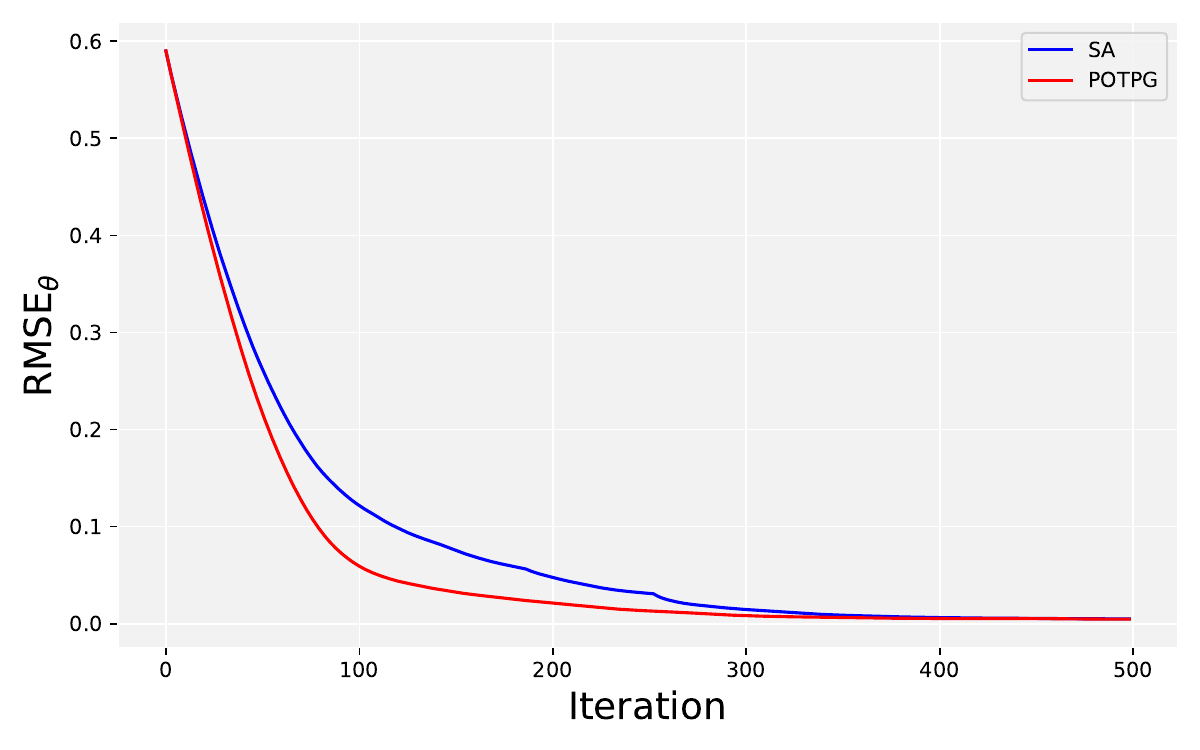}} %
\subfloat[$\text{RMSE}_{\widehat{J}}$, $\xi=0.8$]{\includegraphics[width=0.5\textwidth]{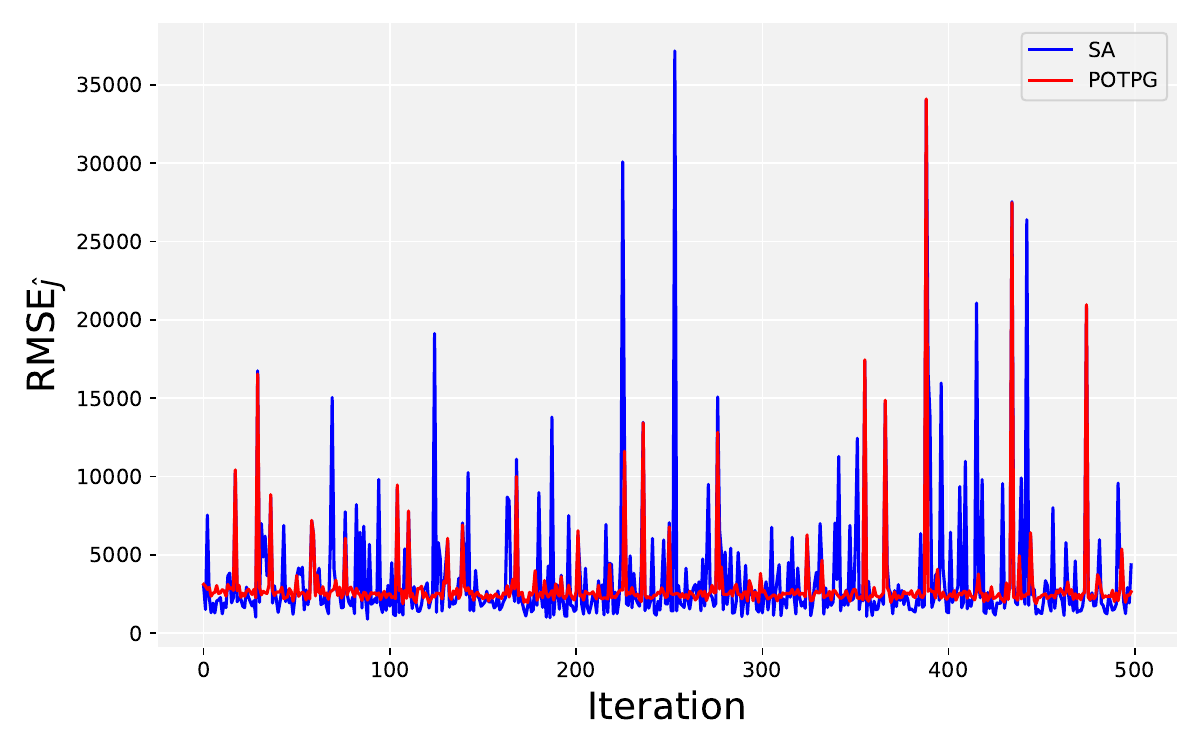}}%
\caption{Training performance for the POTPG algorithm and the sample averaging (SA) benchmark. Left column: RMSE of policy parameter estimate $\text{RMSE}_\theta$. Right column: RMSE of the objective function (the CVaR) $\text{RMSE}_{\widehat{J}}$. RMSE metrics are computed over $R=50$ runs.}
\label{fig:SimulResultsGPD}
\end{figure}

Figure \ref{fig:SimulResultsGPD} reports metrics $\text{RMSE}_\theta$ and $\text{RMSE}_{\widehat{J}}$ with respect to iteration $j=1,\ldots, M$ for the three different values of tail parameter $\xi$. The POTPG outperforms the SA benchmark in all experiments as the RMSE on the optimal policy parameter decreases faster for the former approach. The extent of out-performance increases when the tail thickness (i.e.~parameter $\xi$) increases. This is because sample averaging relies on only four observations, i.e.~$n(1-\alpha)=4$, coming from the tail of distribution, which is increasingly unstable as tail thickness increases. The POT approach better alleviates this issue by using many more observations from the body of the distribution to extrapolate tail behavior. Moreover, even when having converged to the optimal policy, both methods (POTPG and SA) exhibit residual estimation error in the objective function (i.e.~cumulative discounted costs) CVaR estimate. Though $RMSE_{\hat{J}}$ is generally smaller for the POTPG approach, such methods out-perform SA significantly for the thicker tail case $\xi=0.8$. In conclusion, the thicker the tail of the costs distribution is, the more useful the POTPG approach is.

\section{Application to financial hedging}\label{se:hedging}

We present the application of the POTPG algorithm to a financial risk management problem, namely the dynamic Delta-Gamma hedging of an option. 
The problem of finding the optimal proportion of the Gamma to neutralize when options are very expensive is discussed.

\subsection{The hedging framework}

Time elapsed between consecutive time points are assumed to be weeks (period of length $1/52$ year). The periodic continuously compounded interest rate is $r=0.02/52$.
With $S_0=1,\!000$, let $S_t$ denote the time-$t$ price of a non-dividend-paying stock, whose dynamics is assumed to be a discrete-time version of an exponential normal-inverse Gaussian (NIG) Lévy process: $S_t = S_0 e^{\sum_{m=1}^{t} Z_m}$, with $\{Z_t\}^T_{t=1}$ being, under the physical measure $\mathbb{P}$, i.i.d. random variables with a NIG$(\mathtt{a}^\mathbb{P},\beta^\mathbb{P},\delta^\mathbb{P},\mu^\mathbb{P})$ distribution whose PDF is given by
\begin{equation}
    \phi^{NIG}(x; \mathtt{a}, \beta, \mu, \delta) = \frac{\mathtt{a} \delta e^{\delta \gamma}}{\pi} \frac{K_1 (\mathtt{a} \sqrt{\delta^2 + (x-\mu)^2 }}{\sqrt{\delta^2 + (x-\mu^2 )}} e^{{\beta(x-\mu)}}, \quad x \in \mathbb{R},
\end{equation}
where $K_{\lambda}(x)$ represents the modified Bessel function of the second kind with index $\lambda$, defined as:
\begin{equation}
    K_{\lambda}(x) = \frac{1}{2}  \int_{0}^{\infty} u^{\lambda - 1} e^{-\frac{1}{2}x(u^{-1} + u)} du, \quad x>0.
\end{equation}
Such distribution is known to exhibit fat tails and is therefore well-suited to study the extreme risk minimization framework of this study.

Parameters considered are taken from \cite{godin2016minimizing}, namely $\mathtt{a}^\mathbb{P} = 35.7$, $\beta^\mathbb{P} = -10.8$, $\delta^\mathbb{P}=2.04 \times 10^{-2}$ and $\mu^\mathbb{P} = 6.7 \times 10^{-3}$.

We consider a market with high volatility risk premium where options are costly; as such we assume risk-neutral parameters and identical to the physical ones, except for the delta parameter driving the returns variance, which is inflated by a factor of 4: $\mathtt{a}^\mathbb{Q} = \mathtt{a}^\mathbb{P}$, $\beta^\mathbb{Q} = \beta^\mathbb{P}$, $\delta^\mathbb{Q}=4\delta^\mathbb{P}$ and $\mu^\mathbb{Q} = \mu^\mathbb{P}$. In such a market, fully neutralizing the gamma of the option being hedged is most likely sub-optimal, due to high option cost, and thus determining the best hedge ratio yielding the optimal cost versus risk reduction tradeoff is a non-trivial endeavor which is the problem considered in this section.
 
We assume than any European call option on such stock is priced according to the formula provided in \cite{godin2012contingent} which is based on the mean-correcting martingale measure described in \cite{schoutens2003levy}.
The time-$t$ price of a European call option with strike $E$ providing the time $t'$ payoff $\max(0,S_{t'}-E)$ is
\begin{align}\label{NIG call price}
\Pi(t,\tau, E) &= S_t \left(1 - \Phi^{NIG}\left(\ln\left(\frac{E}{S_t}\right); \ \mathtt{a}^\mathbb{Q}, \ \beta^\mathbb{Q} + 1, \ \delta^\mathbb{Q} \tau, \ [\mu^\mathbb{Q} + \zeta^\mathbb{Q}] \tau \right)\right) \nonumber \\
&\quad - E e^{-r\tau} \left(1 - \Phi^{NIG}\left(\ln\left(\frac{E}{S_t}\right); \ \mathtt{a}^\mathbb{Q}, \ \beta^\mathbb{Q}, \ \delta^\mathbb{Q} \tau, \ [\mu^\mathbb{Q} + \zeta^\mathbb{Q}] \tau\right)\right),
\end{align}
with $\tau =t'-t$ weeks, $\zeta^\mathbb{Q}  = r - \mu^\mathbb{Q} + \delta^\mathbb{Q} (\sqrt{(\mathtt{a}^\mathbb{Q})^2 - (\beta^\mathbb{Q}+1)^2} - \sqrt{(\mathtt{a}^\mathbb{Q})^2 - (\beta^\mathbb{Q})^2})$ and $\Phi^{NIG}$ denoting the CDF of the NIG distribution.
It is straightforward to compute the Delta and the Gamma of such options:
\begin{eqnarray*}
\label{nig-delta}
    \Delta(t,\tau, E) &=& \frac{\partial \Pi(t,\tau, E)}{\partial S_t} = 1 -  \Phi^{NIG} \left( \ln \left(\frac{E}{S_t} \right); \  \mathtt{a}^\mathbb{Q}, \ \beta^\mathbb{Q} + 1, \ \delta^\mathbb{Q} \tau, \ [\mu^\mathbb{Q} + \zeta^\mathbb{Q}] \tau \right).
\\     \Gamma(t,\tau, E) &=& \frac{\partial^2 \Pi(t,\tau, E)}{\partial (S_t)^2} = \frac{1}{S_t} \phi^{NIG} \left( \ln \left(\frac{E}{S_t} \right); \  \mathtt{a}^\mathbb{Q}, \ \beta^\mathbb{Q} + 1, \ \delta^\mathbb{Q} \tau, \ [\mu^\mathbb{Q} + \zeta^\mathbb{Q}] \tau \right). \label{nig-gamma}
\end{eqnarray*}

We consider a financial institution (the hedging agent) which holds a short position in a call option with a strike price $E=S_0$ and maturity $T = 0.5 \times 52 = 26$ weeks. Such option is referred to as the target option. To mitigate the risk associated with the uncertainty related to its payoff, a self-financing hedging portfolio is used. At any time point, the portfolio is invested in three hedging assets, namely a risk-free account, the stock and an option on the stock. The time-$t$ value of the hedging portfolio is denoted $V^\theta_t$ (the superscript $\theta$ refers to its dependence on the policy) and evolves according to
\begin{equation*}
    V^\theta_{t+1}= \underbrace{(V^\theta_t - \psi^{(S)}_t S_t - \psi^{(O)}_t H^{beg}_t)}_{ \text{cash investment}} e^{r} + \psi^{(S)}_t (S_{t+1}- S_t) + \psi^{(O)}_t (H^{end}_{t+1}- H^{beg}_t),
\end{equation*}
with $(\psi^{(S)}_t,\psi^{(O)}_t)$ being the respective portfolio positions on time interval $[t,t+1)$ in the stock and an option used for hedging, and $H^{beg}_t$ and $H^{end}_{t+1}$ being the respective time-$t$ and time-$(t+1)$ price of the hedging option purchased at $t$. The positions are thus rebalanced at each period, and option positions are rolled-over, with the hedging options currently in the portfolio being liquidated at the end of the period while new ones are being purchased. At the start of any period $[t,t+1)$, the option considered for purchase is at-the-money (its strike is $S_t$ and its maturity is  $\tau=0.1 \times 52=5.2$, meaning $10$\% of a year).
As such, $H^{beg}_t = \Pi(t,0.1\times 52, S_t)$ and $H^{end}_{t+1} = \Pi(t+1,0.1\times 52-1, S_{t})$. Note that unless $S_t = S_{t+1}$, $H^{end}_{t+1} \neq H^{beg}_{t+1}$ since options being included in the hedging portfolio change on the various periods.
Moreover, $V^\theta_0 = \Pi(0,T, S_0)$ is the option premium that is initially invested in the hedging portfolio.

The optimal policy should characterize the selection of positions $\psi^{(S)}_t,\psi^{(O)}_t$, $t=0,\ldots,T$ to be included in the hedging portfolio. Assume that the agent wants to be fully Delta-neutral, which is obtained with 
\begin{equation*}
    \psi^{(S)}_t = \underbrace{\Delta(t,T-t, S_0)}_{ \text{target option } \Delta} - \psi^{(O)}_t \underbrace{\Delta(t,0.1\times 52, S_t)}_{ \text{hedging option } \Delta}.
\end{equation*}
However, we assume that the agent might prefer not fully neutralizing the Gamma of the target option due to purchases of hedging options being too costly in a market with large volatility risk premium. The agent shall therefore only neutralize a portion $\theta \in (0,1)$, called the hedge ratio, of the target option Gamma. This leads to $\psi^{(O)}_t \Gamma(t,0.1\times 52, S_t) = \theta \Gamma(t,T-t, S_0)$, and thus to $\psi^{(O)}_t = \theta \frac{ \Gamma(t,T-t, S_0)}{ \Gamma(t,0.1\times 52, S_t) }$. 

The objective of the hedging agent is therefore to find the optimal hedge ratio, which is the optimal policy parameter $\theta$.
A single terminal cost is considered for the agent: $C_t = \mathds{1}_{ \{t=T\} } \left( \max(0,S_{T}-E) -V^{\theta}_T\right)$ and no discount factor is considered $\gamma=1$. The agent thus attempts minimizing risk associated with catastrophic hedging shortfalls at maturity: hence consider $\alpha = 0.999$.

\begin{figure}[htbp]
\begin{center}
\includegraphics[scale=0.45]{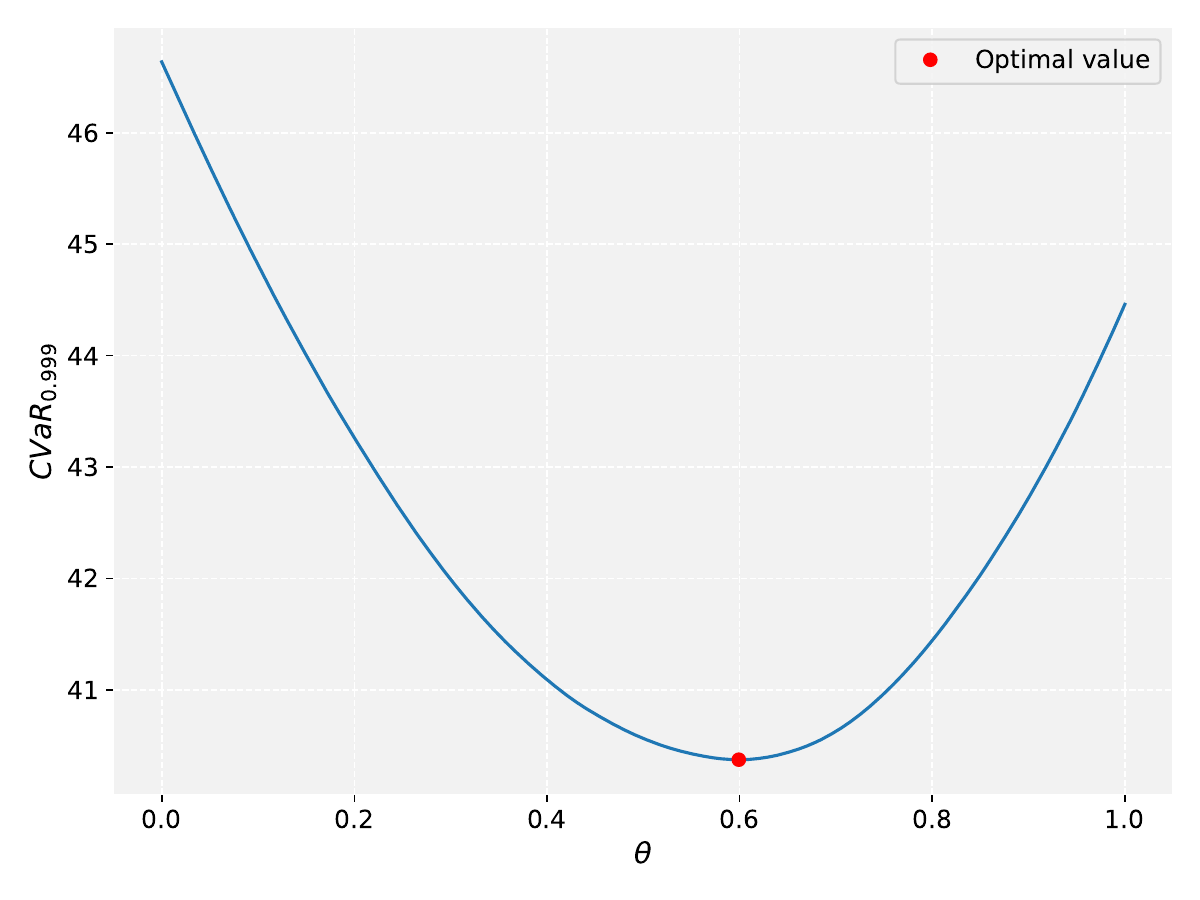} 
\end{center}
\caption{Objective function (CVaR$_{0.999}$ of the hedging shortfall) versus the hedge ratio $\theta$, representing the percentage of the target option Gamma being neutralized. Estimates are obtained by brute force calculations, i.e.~through sample averaing over $1,\!000,\!000$ simulated paths. Red point: optimal value.}
\label{fig: optimal_k}
\end{figure}

Before applying the reinforcement learning procedure, we want to approximate the objective function $J(\theta)$, the CVaR$_{0.999}$ of the hedging shortfall, for various hedge ratios $\theta$. Such approximations are produced with brute force Monte-Carlo simulations, where for several values of $\theta$, $1,\!000,\!000$ realizations of the hedging shortfall $\max(0,S_{T}-E) -V^{\theta}_T$ are produced and sample averaging is applied, i.e.~$J$ is estimated by the $1,\!000$ largest realizations. Figure \ref{fig: optimal_k} reports such estimates, with the optimal hedge ratio being estimated to be $\theta^* = 0.5991$ and the corresponding objective function being $\hat{J}(\theta^*) = 40.37$. 

\vspace{-10pt}
Now apply the POTPG algorithm to the policy optimization problem, and compare its performance to that of the sample averaging method (SA). Such methods are applied with either $n=1,\!000$ or $n=10,\!000$ simulated paths of weekly stock returns.
$R=100$ independent runs are conducted, each comprised of $M=500$ iterations for the case $n=1,\!000$, or $M=150$ iterations when $n=10,\!000$. In each run, the initial policy is set to $\theta^{(0)}=0$. The finite difference shock is $\epsilon=0.05$. The method of moments is used to estimate tail parameters $\xi,\sigma$ in the POTPG algorithms since such method exhibited (in unreported tests) greater stability than maximum likelihood estimates in the presented framework.

\begin{figure}[htp]
    \centering
    \subfloat[RMSE$_\theta$, $n=1,\!000$]{\includegraphics[width=0.45\textwidth]{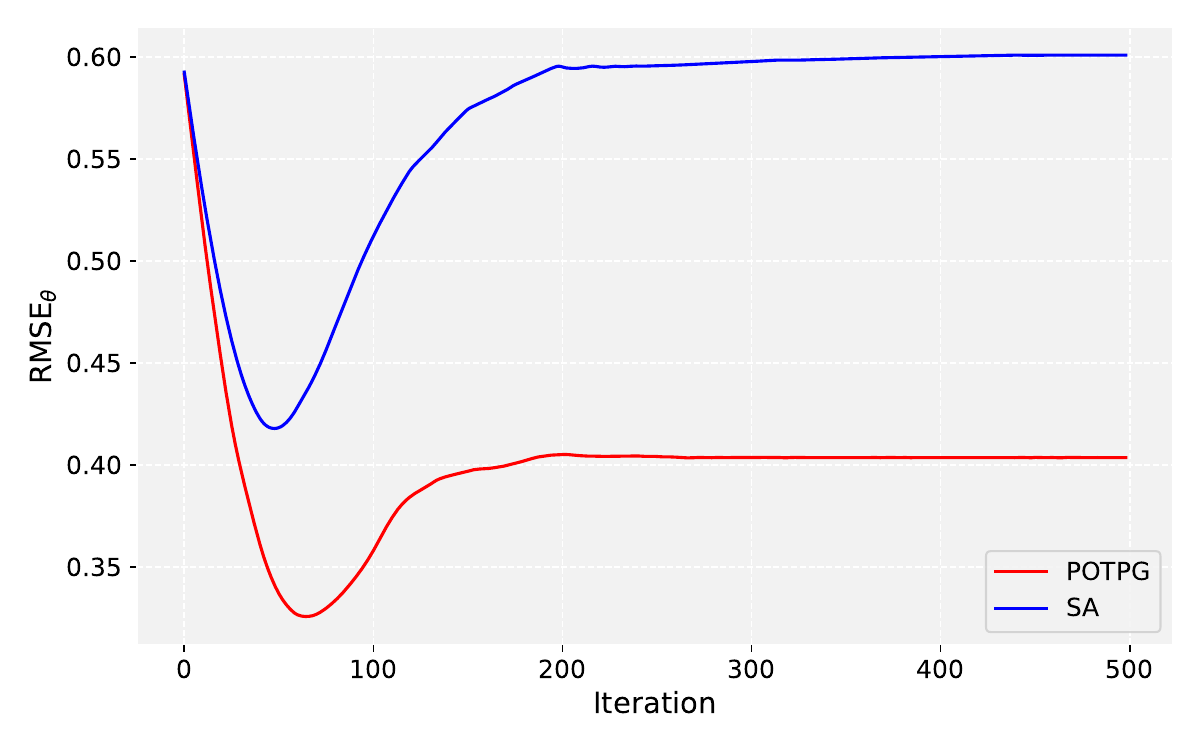}} 
    \subfloat[$\text{RMSE}_{\hat{J}}$, $n=1,\!000$]{\includegraphics[width=0.45\textwidth]{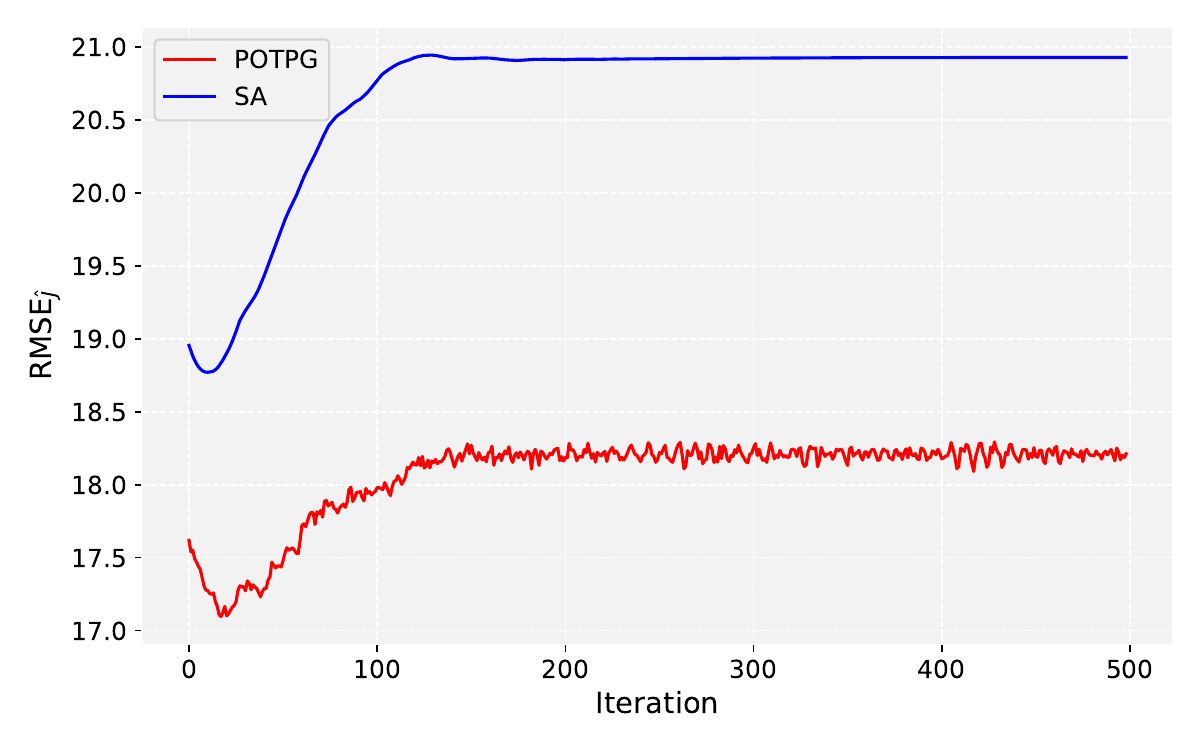}} \\
    \centering
   \centering
    \subfloat[RMSE$_\theta$, $n=10,\!000$]{\includegraphics[width=0.45\textwidth]{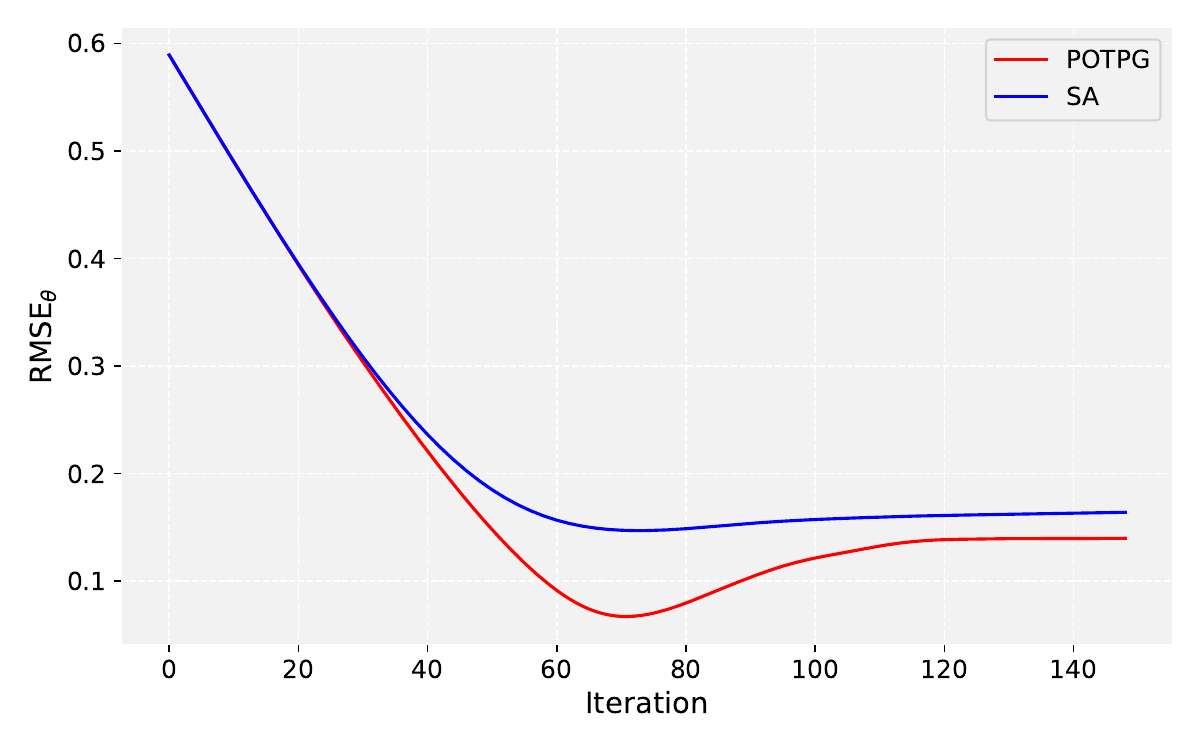}} 
   \subfloat[RMSE$_{\hat{J}}$, $n=10,\!000$]{\includegraphics[width=0.45\textwidth]{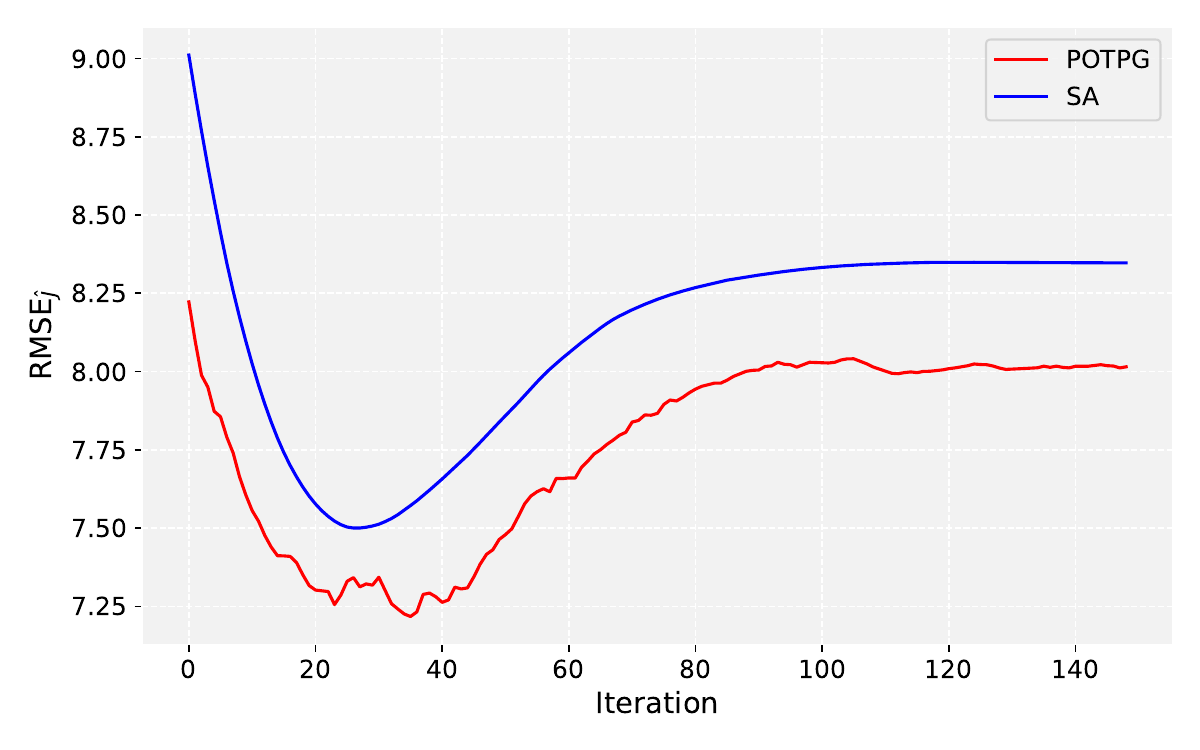}} 
    \caption{Evolution of the RMSE of the estimate of the optimal policy parameter (RMSE$_\theta$) and the corresponding objective function (RMSE$_{\hat{J}}$) over iterations of the POTPG algorithm and the sample averaging (SA) benchmark. Top row: sample size $n=1,\!000$. Bottow row: $n=10,\!000$. Left panels: RMSE$_\theta$. Right panels: RMSE$_{\hat{J}}$.}
    \label{fig: rmse}
\end{figure}

Figure \ref{fig: rmse} reports the performance of the POTPG and SA policy gradient algorithms for the hedging problem, by displaying the evolution of the RMSE (across runs) of the estimate of the optimal policy parameter (RMSE$_\theta$) and the corresponding objective function (RMSE$_{\hat{J}}$) versus the number of iteration conducted. The POTPG algorithm exhibits materially superior performance by exhibiting much lower errors on estimates for the optimal policy parameter and objective function.
The gap in performance between the POTPG and the benchmark is greater for the lower sample size $n=1,\!000$, which highlights that our method has more added value in the context of more severe distribution tail data scarcity.
Note that none of the two methods have the estimated policy parameter converge to the true optimal value (i.e.~RMSE$_\theta$ does not converge to zero), which can be explained by the fact that both methods are biased in finite sample $n$. Nevertheless, we see that higher sample size $n$ increases the precision, with lower RMSEs for the estimates of the policy parameter $\theta^*$ and of the objective function $J$.

\section{Conclusion}\label{se:conclusion}

We propose a policy gradient algorithm based on estimators of tail risk borrowed from extreme value theory to tackle the difficult task of catastrophic risk minimization within a sequential decision making framework. The peaks-over-threshold procedure is used to estimate the CVaR of cumulative costs by leveraging the asymptotic convergence of the tail distribution to a generalized Pareto distribution. We have shown in several simulation experiments, including an application to financial options hedging, that our method can outperform conventional benchmarks relying on the empirical distribution of the cumulative costs. Indeed, such benchmarks can perform quite poorly to mitigate extreme risk when observations in the tail are scarce.

Our method relies on finite difference approximations for the gradient, and as such it work for low-dimensional policies relying on a small number of parameters. An extension of our approach could consist in developing a high-dimensional EVT-based policy gradient framework to tackle more complex problems. This would for instance allow using policies represented by deep neural networks and combine the EVT-based policy gradient with deep reinforcement learning.


\section*{Appendix A $\,\,\,$ The automated threshold selection procedure}
\label{App:Bader}

The automated threshold selection procedure involves testing several thresholds $u_i = \hat{F}^{-1}(q_i)$, $i=1,\ldots,l$, which we choose to be quantiles of pre-determined levels $q_1,\ldots, q_\ell$ of the empirical distribution of the sample $X_1,\ldots,X_n$. For each $i$, denote by $k_i$ the number of threshold excesses. Assuming that a threshold $u_i$ leads to GPD parameter estimates $(\hat{\xi}_{u_i}, \hat{\sigma}_{u_i})$ for the distribution of the excesses $\mathcal{Y}_{u_i}=\{X_i-u_i : X_i>u_i\}$, the Anderson-Darling test statistic for such threshold is
\begin{equation} \label{ad_statistic}
A_{i}^{2}=-k_i-\frac{1}{k_i} \sum_{j=1}^{k_i}(2 j-1)\left[\log \left(Z_{(j,i)}\right)+\log \left(1-Z_{(k_i+1-j,i)}\right)\right].
\end{equation}
where $Z_{(j,i)} = G_{\hat{\xi}_{u_i}, \hat{\sigma}_{u_i}}(Y_{(j,i)})$ with $Y_{(j,i)}$ being the $j^{th}$ smallest excess value, i.e.~among values in $\mathcal{Y}_{u_i}$. The automated selection procedure attempts using the smallest possible threshold for which no threshold above would be deemed inadequate. In the application, we choose $l=20$, $q_1=0.79$, $q_2=0.80,\ldots,q_{20}=0.98$ and $\xi_{max} = 0.9$. 

\begin{remark}
    If the automated threshold selection procedure is unsuccessful, i.e.~$I=\emptyset$, the sample averaging estimate in \eqref{eq: SA} is used for CVaR, as a fallback estimate. 
\end{remark}



\vspace{-5pt}
\begin{algorithm}[H]
    \caption{Automated threshold selection procedure (from \cite{troop2021bias})}
    \label{algo:auto_thresh}
    \begin{algorithmic} 
        \Require Significance parameter $\gamma$, cutoff $\xi_{\text{max}} < 1$, i.i.d. sample $X_1, \ldots, X_n$, threshold percentiles $0 < q_1, \ldots, q_l < 1$.

        \State $I \leftarrow \emptyset$ 
        \For{$i=1,\ldots,l$}
            \State Set $u_i = \hat{F}^{-1}(q_i)$
            \State Compute $(\hat{\xi}_{u_i}, \hat{\sigma}_{u_i})$ from the $k_i$ threshold excesses over $u_i$
            \If{$\hat{\xi}_{u_i} \leq \xi_{\text{max}}$}
                \State Compute $A_i^2$ using \eqref{ad_statistic}
                \State Set $p_i$ to $p$-value for $A_i^2$ using a lookup table
                \State $I \leftarrow I \cup \{i\}$
            \EndIf
        \EndFor

        \If{$I \neq \emptyset$}
            \State Set $W = \{w \in I \, \vert \, -\frac{1}{w} \sum_{i=1}^{w} \log (1-p_{i}) \leq \gamma\}$            
            \If{$W \neq \emptyset$}
                \State Compute $\hat{w}_F = \max W$ 
                \If{$\hat{w}_F = \max(I)$}
                    \State $v \leftarrow \max(I)$
                \Else
                    \State $v \leftarrow \min\{w \in I \, \vert \, w > \hat{w}_F\}$
                \EndIf
            \Else
                \State $v \leftarrow \min(I)$
            \EndIf
            $u \leftarrow u_v$
            \State Return $u_v$, $(\hat{\xi}_{u_v}, \hat{\sigma}_{u_v})$
        \EndIf

    \end{algorithmic}
\end{algorithm}
\vspace{-5pt}

\vspace{-10pt}
\titleformat{\section}{\normalfont\large\bfseries} {\thesection}{1em}{#1}
\bibliographystyle{apalike}
\bibliography{references}  


\end{document}